\documentclass{ifacconf}

\usepackage{graphicx}      % include this line if your document contains figures
\usepackage{natbib}        % required for bibliography
\usepackage{amsmath}
\usepackage{amssymb} 
\usepackage{xcolor}
\usepackage{siunitx}

\usepackage{booktabs}
\usepackage{multirow}
\usepackage{multicol}
\usepackage{caption}
\usepackage[table]{xcolor} 
\usepackage{makecell}

\usepackage{acronym}
\acrodef{uav}[UAV]{Unmanned Aerial Vehicle}	
\acrodef{bo}[BO]{Bayesian Optimization}	
\acrodef{mtbo}[MTBO]{Multi-Task Bayesian Optimization}	
\acrodef{gp}[GP]{Gaussian Process}	
\acrodef{mtgp}[MTGP]{Multi-Task Gaussian Process}	

\begin{document}
\begin{frontmatter}

\title{Multi-Task Bayesian Optimization for Tuning Decentralized Trajectory Generation in Multi-UAV Systems} 

\author[First]{Marta Manzoni} 
\author[First]{Alessandro Nazzari} 
\author[First]{Roberto Rubinacci}
\author[First]{Marco Lovera}

\address[First]{Dipartimento di Scienze e Tecnologie Aerospaziali, Politecnico di Milano, Via La Masa 34, 20156, Milano, Italy (e-mail:  {marta.manzoni, alessandro.nazzari, marco.lovera, roberto.rubinacci}@polimi.it).}

\begin{abstract} 
% Abstract of 50--100 words

This paper investigates the use of Multi-Task Bayesian Optimization for tuning decentralized trajectory generation algorithms in multi-drone systems. We treat each task as a trajectory generation scenario defined by a specific number of drone-to-drone interactions. To model relationships across scenarios, we employ Multi-Task Gaussian Processes, which capture shared structure across tasks and enable efficient information transfer during optimization. We compare two strategies: optimizing the average mission time across all tasks and optimizing each task individually. Through a comprehensive simulation campaign, we show that single-task optimization leads to progressively shorter mission times as swarm size grows, but requires significantly more optimization time than the average-task approach.
\end{abstract}


\end{frontmatter}
%===============================================================================

\section{Introduction}

In recent years, research efforts and real-world applications of Unmanned Aerial Vehicles (UAVs) have increasingly shifted from single-agent to multi-agent systems. This shift is motivated by the expanded capabilities offered by multi-drone systems, including wide-area search, rapid environmental mapping, cooperative inspection, and resilient mission execution. However, realizing these advantages introduces substantial coordination challenges \citep{nazzari2025}. A foundational requirement is the ability to generate safe, feasible motions for multiple UAVs operating simultaneously. This capability is typically provided by decentralized trajectory generation algorithms \citep{rubinacci2025anytime, tordesillas2022}, which solve a trajectory optimization problem in a receding-horizon fashion. Despite significant progress in algorithmic design, the parameters governing these algorithms are almost always hand-tuned or optimized for a single scenario. This approach rarely generalizes: parameters that perform well in one environment may degrade performance, or even cause mission failure in another. 

In this paper, we propose a systematic approach for tuning these parameters by treating the trajectory generation pipeline as a black-box system and leveraging the  Bayesian Optimization (BO) framework \citep{mockus2005bayesian, shahriari2015taking}, as has been successfully applied in recent controller-tuning applications \citep{berkenkamp2016safe}. A central challenge in tuning decentralized trajectory generation algorithms is that their performance is scenario-dependent, and defining what constitutes a "scenario" is itself nontrivial. Factors such as mission goals and UAV density can significantly influence algorithm behavior. To obtain a simple yet informative characterization, we define a scenario based on the number of drone-to-drone interactions occurring during the mission. We then optimize the design parameters using Multi-Task Bayesian Optimization (MTBO), which exploits performance correlations across scenarios with different interaction levels.

The rest of the paper is organized as follows. Section~\ref{sec:MTBO} provides an overview of Gaussian Processes (GPs) and MTBO. Section~\ref{sec:ATOMICA} describes the decentralized trajectory generation algorithm employed for multi-drone coordination. Section~\ref{sec:BOTANICA} presents our multi-task optimization framework. Section~\ref{sec:simulations} reports the numerical simulations conducted to assess the performance of the proposed approach. Finally, Section~\ref{sec:conclusions} summarizes the main findings and concludes the paper.

\section{Multi-task Bayesian Optimization}
\label{sec:MTBO}

This section introduces the theoretical background of MTBO. We first review GPs and their multi-task extensions, and then present the MTBO formulation.

\subsection{Gaussian Processes}

A GP is a flexible, non-parametric probabilistic model for representing unknown functions \citep{williams2006gaussian}. Formally, a GP defines a distribution over functions such that any finite collection of function values follows a joint Gaussian distribution. It is fully specified by a mean function $\mu(\theta): \Theta \rightarrow \mathbb{R}$ and a positive definite covariance, or kernel function, $k(\theta, \theta'): \Theta \times \Theta \rightarrow \mathbb{R}$, which encodes the correlation between pairs of inputs. Let $J(\theta): \Theta \rightarrow \mathbb{R}$ denote the function to be modeled, where $\Theta \subseteq \mathbb{R}^d$ represents the input space of dimension $d$. The GP prior is specified as $J(\theta) \sim \mathcal{GP}(\mu(\theta), k(\theta, \theta'))$, where $\mu(\theta) = \mathbb{E}[J(\theta)]$ is the prior mean and $k(\theta, \theta') = \mathbb{E}[(J(\theta) - \mu(\theta))(J(\theta') - \mu(\theta'))]$ is the prior covariance.

Once a GP prior is defined, closed-form posterior inference is possible given a set of observations. Observations are assumed to be noisy realizations of the true function, modeled as $\hat{J}(\theta_i) = J(\theta_i) + \nu$, for $i = 1, \ldots, n$, where $\nu \sim \mathcal{N}(0, \sigma_\nu^2)$ is Gaussian noise. Given a dataset of $n$ observations $\mathcal{D}_n = \{ (\theta_i, \hat{J}(\theta_i)) \}_{i=1}^{n}$, the posterior mean and covariance are
\\
\begin{align}
\mu_n(\theta) &= \mu(\theta) + k_n(\theta)^\top (K_n + \sigma_\nu^2 I_n)^{-1} \hat{J}_n \label{eq:gp_mean} \\
\Sigma_n(\theta,\theta') &= k(\theta, \theta') - k_n(\theta)^\top (K_n + \sigma_\nu^2 I_n)^{-1} k_n(\theta'), \label{eq:gp_variance}
\end{align}
\\
where $k_n(\theta) = [k(\theta, \theta_1), \ldots, k(\theta, \theta_n)]^\top$ contains the covariances between the new point $\theta$ and all observed inputs, $\hat{J}_n = [\hat{J}(\theta_1), \ldots, \hat{J}(\theta_n)]^\top$ stores the observed values, $K_n \in \mathbb{R}^{n \times n}$ is the kernel matrix with entries $[K_n]_{(i,j)} = k(\theta_i, \theta_j)$, and $I_n$ is the identity matrix of size $n$. 

\subsection{Multi-Task Gaussian Processes}

Multi-Task Gaussian Processes (MTGPs) extend standard GPs to jointly model multiple related functions \citep{bonilla2007multi}. Consider $t = 1, \dots, T$ tasks, each associated with an unknown function $J_t(\theta)$. Instead of fitting independent GPs to each task, a MTGP captures correlations across tasks and inputs using a unified probabilistic framework. 

To achieve this, a MTGP defines a covariance function over input–task pairs $(\theta, t)$, commonly expressed as:
\\
\begin{equation}
\label{eq:MT_covariance}
k_{\text{multi}}((\theta, t), (\theta', t')) = k_t(t, t') \otimes k_\theta(\theta, \theta'),
\end{equation}
\\
where $k_\theta$ models the covariance between inputs and $k_t$ encodes relationships among tasks, and $\otimes$ denotes the Kronecker product. A widely used formulation for $k_t$ is the intrinsic model of coregionalization (ICM) \citep{bonilla2007multi}, which assumes that each task is a linear combination of shared latent functions governed by a common GP prior. Once the kernel structure is defined, inference and prediction proceed analogously to the single-task case. By exploiting correlations across tasks, MTGPs improve data efficiency and enable effective knowledge transfer between related learning problems.

\subsection{Multi-Task Bayesian Optimization}

Bayesian Optimization (BO) \citep{mockus2005bayesian, shahriari2015taking} is a sample-efficient probabilistic framework for the global optimization of an unknown and expensive-to-evaluate function $J(\theta)$. The problem is formulated as:
\\
\begin{equation}
\max_{\theta \in \Theta} J(\theta). 
\end{equation}
\\
BO builds a probabilistic surrogate, typically a GP, to approximate the objective function and to guide the selection of new evaluation points through an acquisition function that balances exploration and exploitation. At iteration $n$, the next query is selected as:
\begin{equation}
\theta_n = \operatorname*{\arg\max}_{\theta \in \Theta} \alpha_n(\theta),
\end{equation}
where popular choices for $\alpha(\theta)$ include Expected Improvement (EI) \citep{jones1998efficient}, Probability of Improvement, or Upper Confidence Bound (UCB) \citep{srinivas2012information}. After evaluating $\hat{J}(\theta_n)$, the surrogate model is updated, and the process continues until convergence or until a maximum number of iterations $N_{\text{max}}$ is reached.

MTBO \citep{swersky2013multi, dai2020multi} extends this framework to the optimization of multiple related tasks. Instead of running independent BO processes for each task, MTBO employs a MTGP to jointly model all task-specific objective functions. The MTGP uses the kernel structure in \eqref{eq:MT_covariance}, which captures correlations across both inputs and tasks. Through the inter-task covariance function $k_t$, observations gathered from any task contribute to the learning of all other tasks. This shared modeling structure enables efficient transfer of information across tasks, often leading to faster convergence and improved sample efficiency compared to single-task BO, especially when tasks are strongly correlated.

\section{Decentralized multi-UAV trajectory optimization: ATOMICA}
\label{sec:ATOMICA}

The goal of a multi-drone trajectory generation algorithm is to guide each UAV from its start position to a specified goal while avoiding collisions with other UAVs. To ensure scalability, this problem is typically solved in a decentralized manner, where each UAV treats the other agents as dynamic obstacles and coordinates through inter-agent communication. In this work, we employ the recently developed receding-horizon anytime algorithm \textsc{atomica} \citep{rubinacci2025anytime}.

The trajectory of each UAV is parameterized using piecewise polynomials:
\begin{equation}
    p_i(t) = 
    \begin{cases}
        \sum_{k=0}^{n} B_k^n(t) P_{i,k}^{(1)}, & \text{if } t_0 \leq t < t_1, \\
        \vdots & \vdots \\
        \sum_{k=0}^{n} B_k^n(t) P_{i,k}^{(m)}, & \text{if } t_{m-1} \leq t \leq t_m, 
    \end{cases}
\end{equation}

where $m$ is the number of segments, $B_k^n(u_l) = \binom{n}{k} u_l^k (1 - u_l)^{n - k}, u_l=\frac{t-t_l}{t_{l+1}-t_l}\in [0,1]$ are the Bernstein basis functions, and $P_{i,k}^{(l)} \in \mathbb{R}^3,  k=0,\ldots,n$, are the control points for UAV $i$ in the $l^{th}$ segment. 

At each planning step, each UAV \(i\) solves the following trajectory-optimization problem:

\begin{equation}
\begin{aligned}
\label{eq:uav_problem_formulation}
    & \underset{p_i(t)}{\text{minimize}} & & \rho \|p_i(T_i) - p_{ig} \|_2 +  \int_{t_0}^{t_0+T_i} \| \ddddot{p}_i(t) \|_2^2 \\
    & \text{subject to} & & \|p_i(t)-p_j(t) \|_2 \ge d_{ij,min}, \forall t \in [t_0,t_0+T_i], \\ & & & \forall j \neq i, j=1,\ldots,N, \\
    %p_i(t) \in C_{ij}, \quad \forall j \neq i, \ j=1,\ldots,N,  \\
    & & & \|\dot{p}_i(t)\|_{2} \le v_{\text{max}}, \quad \forall t \in [t_0,t_0+T_i], \\ 
    & & & \|\ddot{p}_i(t)\|_{2} \le a_{\text{max}}, \quad \forall t \in [t_0,t_0+T_i], \\
    & & & p_i(t_0) = p_{i0}, \quad \dot{p}(t_0) = v_{i0}, \quad \ddot{p}(t_0) = a_{i0}, \\
\end{aligned}
\end{equation}

where $N$ is the number of agents, $p_i(t)$ is the $3D$ trajectory of the \(i^{th}\) UAV, \(\rho\) is a tuning parameter, \(p_{ig}\) is the goal position, $T_{i}$ is the planning horizon, $v_{\text{max}}$ and $a_{\text{max}}$ are the maximum allowable velocity and acceleration, respectively, and $d_{ij,\text{min}}$ is the minimum safety distance between the \(ij\) UAVs. 

Starting from a feasible initial guess, the optimization problem is solved within a user-specified time budget \(\delta_{opt}\) using the convex-concave procedure as described in \citep{rubinacci2025anytime}. 

\section{Optimization framework}
\label{sec:BOTANICA}

In this section, we present the main contribution of the paper: a systematic framework for tuning the parameters of a decentralized trajectory optimization algorithm.  

\subsection{Parameter selection and optimization objective}

The performance of \textsc{atomica} is strongly influenced by a set of design parameters of the trajectory optimization problem. These parameters are collectively denoted as $\theta = [m, T, \delta_{opt}]$, where:
\begin{itemize}
    \item $m$: the number of polynomial segments used to parameterize each trajectory,
    \item $T$: the planning horizon, assumed to be identical for all UAVs ($T_i=T \quad \forall i$),
    \item $\delta_{opt}$: the maximum computation time allocated to the trajectory optimization step.
\end{itemize}

In a centralized formulation, one would prefer a long planning horizon \(T\) to approximate the behavior of an offline planner, a generous time budget \(\delta_{opt}\) to allow the optimizer to converge, and a sufficiently large number of segments \(m\) to represent a rich class of trajectories. However, when the problem must be solved in real time and in a decentralized manner, these choices become significantly more delicate. Longer horizons require more segments, increasing computational load, while a small \(\delta_{opt}\) limits the achievable solution quality. As a result, selecting parameter values that strike a balance between trajectory quality, responsiveness, and computational feasibility is nontrivial. Poor parameter choices can lead to deadlock situations where multiple UAVs fail to generate collision-free trajectories that guide them to their goals.

To quantify performance under different parameter choices, we use the mission time \(T_m\), defined as the time required by the last UAV to reach its destination. 

\subsection{Scenario and task definition}

A mission scenario is specified by the number of UAVs involved and by their initial and final positions. For simplicity, we consider environments without static obstacles. To obtain a compact and informative measure of scenario difficulty, we characterize each scenario by the number of drone-to-drone interactions. Specifically, we count the number of pairwise intersections between the straight-line paths connecting each agent’s initial and final positions. This interaction count provides a proxy for the level of coordination required: higher values correspond to denser coupling between agents and typically lead to more challenging planning problems.

Performing MTBO over the full space of possible task configurations would be computationally prohibitive. Therefore, the training set must be carefully designed to include the minimal number of tasks that still capture the diversity of operational scenarios the swarm may encounter. To this end, we define a set of fundamental, or ‘base,’ tasks that capture the essential interactions likely to occur among UAVs during a mission. The key idea is that more complex missions can be decomposed into combinations of these base tasks, allowing the framework to generalize to novel or more challenging scenarios.

\begin{figure}
    \centering
    \includegraphics[width=0.9\linewidth]{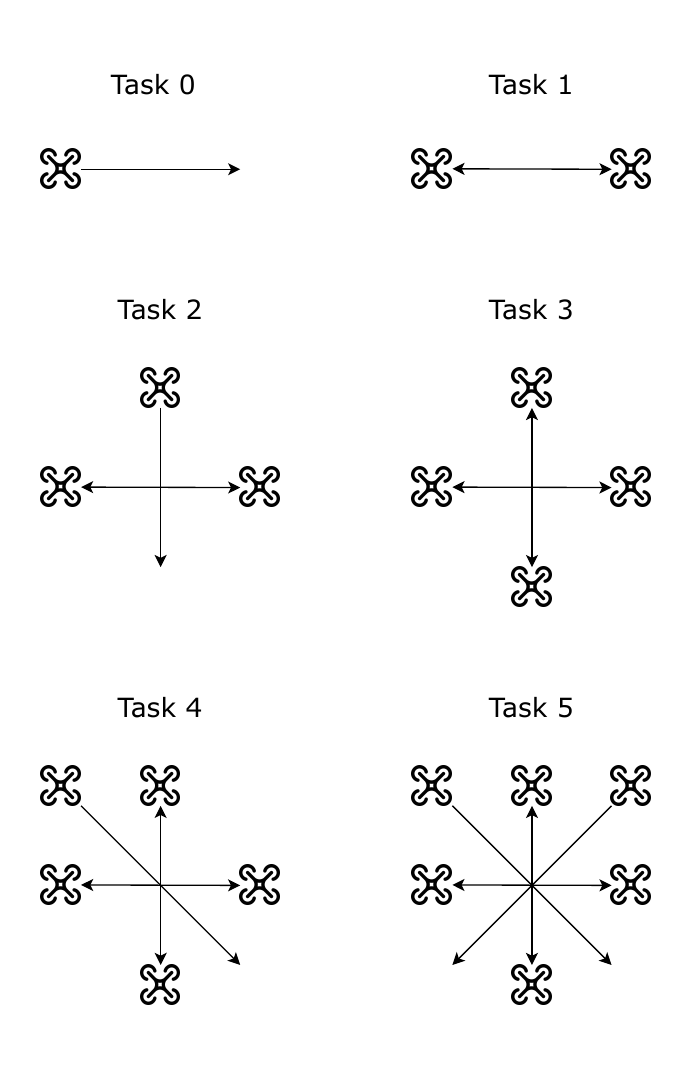}
    \caption{Graphical representation of the six base training tasks used in the optimization framework. Single-headed arrows indicate a UAV that must move toward the arrow tip and stop at that goal position, while double-headed arrows indicate pairs of UAVs that must swap their positions.}
    \label{fig:base_tasks}
\end{figure}

We selected six base tasks, each defined by the number of agents, their initial positions, and their respective goal positions. The task index $t \in \{0, \dots, 5\}$ reflects increasing mission complexity, ranging from a single vehicle flying independently to multi-agent scenarios with progressively more intersecting trajectories. The tasks are illustrated in Figure~\ref{fig:base_tasks}, where single-headed arrows indicate a UAV that must move toward the arrow tip and stop at that goal position, while double-headed arrows indicate pairs of UAVs that must swap their positions.

\subsection{Multi-task parameter optimization}
\label{sec:MT_parameter_optimization}

MTBO is used to optimize the key configuration parameters of \textsc{atomica}. Two formulations are considered. The \textit{average approach} \citep{swersky2013multi} optimizes a single scalar objective across all tasks, yielding a single set of parameters shared by every task. The \textit{single-task approach} \citep{dai2020multi} searches for task-specific optima, producing a distinct set of parameters for each task.

\subsubsection{Average approach.}

The average formulation defines a single objective function representing the mean performance across $T$ related tasks. Each task $t$ has its own objective function $J_t(\theta)$, and the averaged objective is obtained by taking the arithmetic mean of all $T$ task-specific performances. The goal is to find a single set of parameters that is shared across all tasks and maximizes this average performance.

Given $n$ noisy observations $\mathcal{D}_n = \{ ((\theta_i, t_i), \hat{J}_t(\theta_i, t_i)) \}_{i=1}^{n}$, the MTGP posterior provides task-specific predictive means $\mu_n(\theta, t)$ and covariances $\Sigma_n((\theta, t), (\theta', t'))$. The predictive mean and variance of the averaged objective are:
\begin{align}
\Bar{\mu}_n(\theta) &= \frac{1}{T} \sum_{t=1}^{T} \mu_n(\theta, t) \label{eq:avg_mean}\\
\Bar{\sigma}_n^2(\theta) &= \frac{1}{T^2} \sum_{t=1}^{T} \sum_{t'=1}^{T} \Sigma_n((\theta, t), (\theta, t')). \label{eq:avg_variance}
\end{align}

At each iteration, only a single task is actually evaluated. To select it, a two-stage heuristic is used. First, missing observations for the other tasks are imputed using the predictive means from the MTGP. Next, the EI acquisition function \citep{jones1998efficient}, evaluated on the averaged objective, is maximized to identify a promising parameter set $\theta$. Given this $\theta$, the task $t$ that maximizes the single-task EI is then selected for evaluation.

\subsubsection{Single-task approach.}

The single-task formulation \citep{dai2020multi} treats each task individually. The goal is to find a distinct input configuration $\theta_t$ for each task such that the corresponding function $J_t$ reaches its maximum at $\theta_t$. At each iteration $n$, all tasks $\{J_1, J_2, \dots, J_T\}$ are evaluated at their respective query points $\{\theta_1, \theta_2, \dots, \theta_T\}$, which may differ between tasks. The resulting observations are noisy, modeled as $\hat{J}_t(\theta_t) = J_t(\theta_t) + \nu_t$, where $\nu_t \sim \mathcal{N}(0, \sigma_t^2)$ and the noise terms are assumed to be independent across tasks. The set of query points at iteration $n$ is denoted $\theta_n := \{ \theta_{1,n}, \theta_{2,n}, \ldots, \theta_{T,n} \}$, and the collection of all observations up to iteration $n$ is $\mathcal{D}_n = \{ (\theta_{t,i}, \hat{J}_t(\theta_{t,i})) \mid t=1\dots T, i=1\dots n\}$. The posterior of the joint GP conditioned on $\mathcal{D}_n$ is characterized by mean $\mu_n$ and covariance $\Sigma_n$. Task-specific posterior mean and variance are denoted $\mu_{t,n}$ and $\sigma_{t,n}$, respectively. The next set of query points, $\theta_{n+1} := \{ \theta_{1,n+1}, \ldots, \theta_{T,n+1} \}$ is selected using the multi-task GP-UCB acquisition function:
\\
\begin{equation}
\theta_{t,n+1} := \arg\max_{\theta_t} \, \mu_{t,n}(\theta_t) + \sqrt{\beta_{n+1}} \, \sigma_{t,n}(\theta_t), \quad t=1,\dots,T,
\end{equation}
where $\beta_{n+1}$ is an exploration parameter that balances exploitation and exploration.

\subsection{Handling unsuccessful evaluations}

During the learning process, certain parameter configurations $\theta$ may cause the multi-UAV system to fail to complete the mission, for example due to deadlocks. These failures provide valuable information: they indicate regions of the parameter space prone to unsuccessful behavior, guiding the optimization procedure to avoid these areas and focus on configurations that are feasible and likely to yield high performance.

Formally, let the evaluation return a noisy performance measure $\hat{J}_t(\theta)$ and a success indicator $g_t(\theta) \in \{0,1\}$, where $g_t(\theta)=1$ if all UAVs successfully complete the mission and $g_t(\theta)=0$ otherwise. The reward is then defined as
\\
\begin{align}
\tilde{J}_t(\theta) = 
\begin{cases} 
\hat{J}_t(\theta),  \quad &g_t(\theta)=1 \\
\hat{J}_{penalty}, \quad &g_t(\theta)=0,
\end{cases}
\end{align}
\\
where $\hat{J}_{penalty}$ is a fixed heuristic penalty, typically chosen slightly worse than the performance of the worst successful configuration across all tasks. When domain knowledge is unavailable, $\hat{J}_{\mathrm{penalty}}$ can be estimated from the initial design used for hyperparameter tuning. \textcolor{blue}.

\subsection{Implementation details and optimization results}

%\textcolor{red}{Da decidere se lasciare qui o spostare nella sezione dei risultati. Io lascerei qui} 

The implementation choices described here apply to both the average and single-task optimization strategies. The search domain explored by the optimizer is defined by the following bounds:
\\
\begin{equation*}
    m \in \{2,3,4,5,6,7\}, \qquad 
T \in [0.3, 2.0], \qquad 
\delta_{\mathrm{opt}} \in [0.01, 0.2].
\end{equation*}
\\
The MTBO procedures are executed with a maximum budget of $N_{\max}=350$ iterations.

To model the multi-task objective, we assume a zero-mean prior and construct the kernel function in the form of \eqref{eq:MT_covariance}. The input-dependent component \(k_\theta\) is modeled using a Matérn kernel with $\nu=3/2$, a choice well suited for noisy functions that are differentiable but not overly smooth. Automatic relevance determination is employed to assign separate length scales to each input dimension. Correlations between tasks are represented through the ICM, implemented with a coregionalization rank of \(2\). This low-rank structure is sufficient to capture dominant shared latent factors across tasks while avoiding overparameterization of the task covariance matrix. Increasing the rank beyond \(2\) do not yield improvements in marginal likelihood or optimization performance.
The kernel hyperparameters are optimized by maximizing the marginal likelihood with respect to an initial dataset specifically constructed to ensure good coverage of the input–task space. For each task, 18 initial points were generated. Continuous parameters were sampled using Latin Hypercube Sampling, while discrete parameters and tasks were drawn cyclically to ensure uniform coverage. 

The optimized parameters obtained for both the average and single-task approaches are reported in table~\ref{tab:atomica_params}.

\renewcommand{\arraystretch}{1.3}
\begin{table}[h]
    \centering
    \captionsetup{skip=10pt}
    \caption{Optimized \textsc{atomica} parameters from average and single-task approaches.}
    \setlength{\tabcolsep}{6pt}
    \begin{tabular}{l c c c c}
    \toprule
    \textbf{Method} & \textbf{Task} & \multicolumn{3}{c}{\textbf{Parameters}}  \\
    \cmidrule(lr){3-5}
                        &                 & $m$ & $T$ & $\delta_{opt}$ \\
        \midrule
    Average MTBO & -- & 3 & 1.058 & 0.125 \\
    \midrule
    \multirow{6}{*}{Single-task MTBO} 
            & Task 0 & 3 & 1.079 & 0.071 \\
            & Task 1 & 4 & 1.099 & 0.153 \\
            & Task 2 & 4 & 0.776 & 0.072 \\
            & Task 3 & 4 & 0.798 & 0.075 \\
            & Task 4 & 4 & 0.776 & 0.076 \\
            & Task 5 & 4 & 1.175 & 0.025 \\
    \bottomrule
    \end{tabular}
    \label{tab:atomica_params}
\end{table}

\section{Simulation results}
\label{sec:simulations}

The following section presents simulations performed to assess the performance of the proposed optimization framework.

\subsection{Simulation setup}

To assess the performance of our proposed framework, we conducted a set of simulations across different swarm sizes and task complexities. 

In the test scenarios, UAVs are initially placed randomly along the perimeter of a circle with radius \SI{10}{\meter}, with each vehicle tasked to reach a different goal position on the same circumference, also selected at random. We performed tests with swarms of 2, 4, 6, 8, 10, and 12 drones. For each swarm size, we randomly generated three different sets of starting and goal positions. For each configuration, we ran five simulation trials, recording the mission time for each run. We then calculated the average mission time across all five runs and three scenarios. 
We assume that each UAV has perfect knowledge of the initial and goal positions of all other agents. This assumption allows us to isolate the effect of the parameters on the mission time, without additional variability introduced by trajectory-estimation errors.

\subsection{Parameter assignment strategy}

In our evaluation, we compared two sets of parameters: the \textit{average-task parameters} and the \textit{single-task parameters}, listed in table~\ref{tab:atomica_params}.  
When using the average-task parameters in simulation, all UAVs operate with the same parameter set, independent of the number of expected drone-to-drone interactions along their trajectories.
In contrast, when the single-task parameters are used for simulation, each UAV selects the parameter set most appropriate for its expected interaction level. To enable this, each scenario is decomposed into the base tasks previously defined during training. Before the mission starts, every UAV determines how many interactions it will face by analysing the straight-line segments connecting the initial and goal positions of all agents. For each drone, we count how many of these segments intersect with its own. Based on this information, the UAV identifies the corresponding base task and applies the associated optimized parameters.

\begin{figure}
    \centering
    \includegraphics[width=\linewidth]{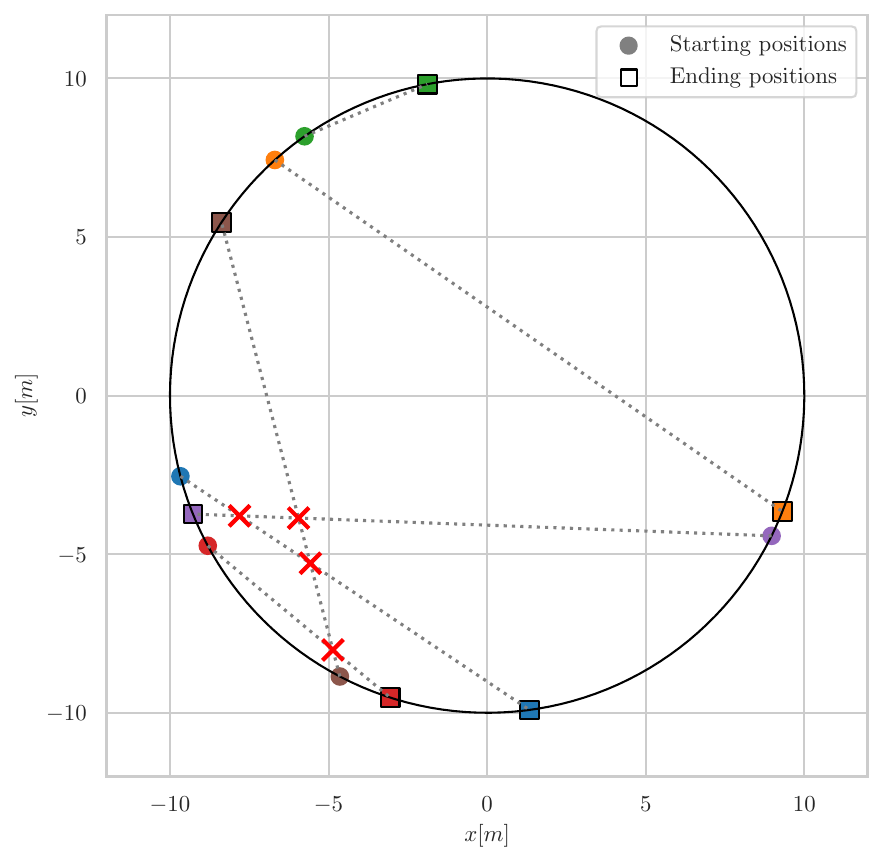}
    \caption{Sample scenario with six UAVs. Circle markers indicate starting positions, and square markers indicate goal positions. Each color corresponds to a different drone. Dashed lines show the straight-line paths connecting start and goal positions, while red crosses mark intersections between paths.}
    \label{fig:scenario}
\end{figure}

Figure~\ref{fig:scenario} illustrates a sample scenario with six UAVs. The dashed lines represent the straight-line segments connecting the initial and goal positions of each drone. The path of the purple UAV intersects with the paths of the blue and brown UAVs, resulting in two expected interactions. Therefore, the purple drone loads the parameters corresponding to Task 3 at the beginning of the mission. In contrast, the orange and green UAVs travel along paths that do not intersect with any other drones, so they use the parameters associated with Task 0. The remaining UAVs select parameters in the same way.

\subsection{Results and discussion}

Table~\ref{tab:mission_times} reports the average mission time $T_m$ computed over the five runs and three randomly generated scenarios for each swarm size and each parameter set. The results reveal a clear performance trend: for small swarms (2–4 UAVs), both the average-task and single-task parameters achieve comparable mission times, with a slight advantage for the average-task parameters. For intermediate swarm sizes (6 UAVs), performance remains similar, but single-task parameters become slightly faster. For larger swarms (8–12 UAVs), single-task parameters consistently achieve shorter mission times, with the performance gap increasing as the swarm size grows. This behavior can be explained by the way the two parameter sets are obtained.

In the average-task approach, a single parameter set is optimized to minimize the average mission time across all base tasks, as detailed in Section~\ref{sec:MT_parameter_optimization}. This naturally results in parameters that perform best for low-density tasks with fewer interactions. In such tasks, reducing mission time has a larger impact on the overall average because trajectories can be more aggressive, and parameter tuning has a stronger effect. Conversely, high-density tasks involve many intersections, which limit trajectory aggressiveness and reduce the benefit of tuning. As a result, the optimization favors parameters that improve performance in low-density scenarios, where their effect on the average objective is greatest. 

The single-task parameters, in contrast, are optimized separately for each base task, allowing them to adapt to the specific number of drone-to-drone interactions in each task.

This explains the performance observed in testing. In test scenarios with small swarms (2–6 UAVs), most drones operate in base tasks with few interactions, so the tasks encountered during testing match those effectively optimized by the average-task approach, resulting in similar performance for both methods. In larger swarms (8–12 UAVs), drones are more likely to encounter base tasks with many intersections, where single-task parameters outperform the average-task set.

Although the single-task approach provides better performance, its optimization requires nearly an order of magnitude more computational time than the average-task approach. Selecting between the two methods, therefore, requires balancing performance gains against computational cost.

\renewcommand{\arraystretch}{1.3}
\begin{table}[h]
    \centering
    \captionsetup{skip=10pt}
    \caption{Mission time $T_m$ (in seconds) for different numbers of UAVs using average and single-task parameters.}
    \setlength{\tabcolsep}{6pt}
    \begin{tabular}{c c c}
    \toprule
    \textbf{Swarm size} & \multicolumn{2}{c}{\textbf{Mission time $T_m$ [s]}}  \\
    \cmidrule(lr){2-3}  & \makecell{average-task \\ params} & \makecell{single-task \\ params} \\
    \midrule
    2 & 10.03 & 10.16 \\
    4 & 11.44 & 11.52 \\
    6 & 11.93 & 11.77 \\
    8 & 12.94 & 12.21 \\
    10 & 14.20 & 13.03 \\
    12 & 14.08 & 11.86 \\
    \bottomrule
    \end{tabular}
    \label{tab:mission_times}
\end{table}

\section{Conclusions}
\label{sec:conclusions}

This work investigated the use of MTBO to tune the parameters of a decentralized trajectory generation algorithm for multi-drone systems. We compared two strategies: the \textit{average approach} and the \textit{single-task approach}. Simulation results indicate that both approaches yield comparable performance for small and medium-sized swarms (2 to 6 UAVs), while the single-task approach consistently outperforms the average approach as swarm size increases. Despite its superior performance, the single-task approach requires significantly more computational resources. Therefore, choosing between the two methods involves a trade-off between computational cost and achievable performance.

%\begin{ack}
%Place acknowledgments here.
%\end{ack}

% \bibliography{ifacconf}             

\begin{thebibliography}{12}
\providecommand{\natexlab}[1]{#1}
\providecommand{\url}[1]{\texttt{#1}}
\providecommand{\urlprefix}{URL }
\expandafter\ifx\csname urlstyle\endcsname\relax
  \providecommand{\doi}[1]{doi:\discretionary{}{}{}#1}\else
  \providecommand{\doi}{doi:\discretionary{}{}{}\begingroup \urlstyle{rm}\Url}\fi

\bibitem[{Berkenkamp et~al.(2016)Berkenkamp, Schoellig, and Krause}]{berkenkamp2016safe}
Berkenkamp, F., Schoellig, A.P., and Krause, A. (2016).
\newblock Safe controller optimization for quadrotors with gaussian processes.
\newblock In \emph{2016 IEEE international conference on robotics and automation (ICRA)}, 491--496. IEEE.

\bibitem[{Bonilla et~al.(2007)Bonilla, Chai, and Williams}]{bonilla2007multi}
Bonilla, E.V., Chai, K., and Williams, C. (2007).
\newblock Multi-task gaussian process prediction.
\newblock \emph{Advances in neural information processing systems}, 20.

\bibitem[{Dai et~al.(2020)Dai, Song, and Yue}]{dai2020multi}
Dai, S., Song, J., and Yue, Y. (2020).
\newblock Multi-task bayesian optimization via gaussian process upper confidence bound.
\newblock In \emph{ICML 2020 workshop on real world experiment design and active learning}, volume~60, 61.

\bibitem[{Jones et~al.(1998)Jones, Schonlau, and Welch}]{jones1998efficient}
Jones, D.R., Schonlau, M., and Welch, W.J. (1998).
\newblock Efficient global optimization of expensive black-box functions.
\newblock \emph{Journal of Global optimization}, 13(4), 455--492.

\bibitem[{Mockus(2005)}]{mockus2005bayesian}
Mockus, J. (2005).
\newblock The bayesian approach to global optimization.
\newblock In \emph{System Modeling and Optimization: Proceedings of the 10th IFIP Conference New York City, USA, August 31--September 4, 1981}, 473--481. Springer.

\bibitem[{Nazzari et~al.(2025)Nazzari, Rubinacci, and Lovera}]{nazzari2025}
Nazzari, A., Rubinacci, R., and Lovera, M. (2025).
\newblock Tacos: Task agnostic coordinator of a multi-drone system.
\newblock \urlprefix\url{https://arxiv.org/abs/2510.01869}.

\bibitem[{Rubinacci et~al.(2025)Rubinacci, Nazzari, and Lovera}]{rubinacci2025anytime}
Rubinacci, R., Nazzari, A., and Lovera, M. (2025).
\newblock Anytime trajectory optimization for multi-drone systems with guaranteed collision avoidance.
\newblock \emph{IEEE Control Systems Letters}.

\bibitem[{Shahriari et~al.(2015)Shahriari, Swersky, Wang, Adams, and De~Freitas}]{shahriari2015taking}
Shahriari, B., Swersky, K., Wang, Z., Adams, R.P., and De~Freitas, N. (2015).
\newblock Taking the human out of the loop: A review of bayesian optimization.
\newblock \emph{Proceedings of the IEEE}, 104(1), 148--175.

\bibitem[{Srinivas et~al.(2012)Srinivas, Krause, Kakade, and Seeger}]{srinivas2012information}
Srinivas, N., Krause, A., Kakade, S.M., and Seeger, M.W. (2012).
\newblock Information-theoretic regret bounds for gaussian process optimization in the bandit setting.
\newblock \emph{IEEE transactions on information theory}, 58(5), 3250--3265.

\bibitem[{Swersky et~al.(2013)Swersky, Snoek, and Adams}]{swersky2013multi}
Swersky, K., Snoek, J., and Adams, R.P. (2013).
\newblock Multi-task bayesian optimization.
\newblock \emph{Advances in neural information processing systems}, 26.

\bibitem[{Tordesillas and How(2022)}]{tordesillas2022}
Tordesillas, J. and How, J.P. (2022).
\newblock Mader: Trajectory planner in multiagent and dynamic environments.
\newblock \emph{IEEE Transactions on Robotics}, 38(1), 463--476.
\newblock \doi{10.1109/TRO.2021.3080235}.

\bibitem[{Williams and Rasmussen(2006)}]{williams2006gaussian}
Williams, C.K. and Rasmussen, C.E. (2006).
\newblock \emph{Gaussian processes for machine learning}, volume~2.
\newblock MIT press Cambridge, MA.

\end{thebibliography}
% bib file to produce the bibliography
% with bibtex (preferred)

\end{document}